\documentclass[10pt]{article} %
\usepackage[preprint]{tmlr}

\usepackage{amsmath,amsfonts,bm}

\def\eqref#1{equation~\ref{#1}}

\def\1{\bm{1}}

\DeclareMathAlphabet{\mathsfit}{\encodingdefault}{\sfdefault}{m}{sl}
\SetMathAlphabet{\mathsfit}{bold}{\encodingdefault}{\sfdefault}{bx}{n}

\usepackage{hyperref}
\usepackage{url}
\usepackage{comment}
\usepackage{listings}

\includecomment{workedexample}

\usepackage{natbib} %

\usepackage{microtype}
\usepackage{graphicx}
\usepackage{caption}
\usepackage{subcaption}
\usepackage{booktabs} %
\usepackage{longtable}
\usepackage{bbm}

\usepackage[capitalize,noabbrev]{cleveref}

\definecolor{solarized@base03}{HTML}{002B36}
\definecolor{solarized@base02}{HTML}{073642}
\definecolor{solarized@base01}{HTML}{586e75}
\definecolor{solarized@base00}{HTML}{657b83}
\definecolor{solarized@base0}{HTML}{839496}
\definecolor{solarized@base1}{HTML}{93a1a1}
\definecolor{solarized@base2}{HTML}{EEE8D5}
\definecolor{solarized@base3}{HTML}{FDF6E3}
\definecolor{solarized@yellow}{HTML}{B58900}
\definecolor{solarized@orange}{HTML}{CB4B16}
\definecolor{solarized@red}{HTML}{DC322F}
\definecolor{solarized@magenta}{HTML}{D33682}
\definecolor{solarized@violet}{HTML}{6C71C4}
\definecolor{solarized@blue}{HTML}{268BD2}
\definecolor{solarized@cyan}{HTML}{2AA198}
\definecolor{solarized@green}{HTML}{859900}

\hypersetup{
    colorlinks,
    linkcolor={solarized@orange},
    citecolor={solarized@yellow},
    urlcolor={solarized@blue}
}

\title{Does `Deep Learning on a Data Diet' reproduce?\\ Overall yes, but GraNd at Initialization does not}

\author{\name Andreas Kirsch \email andreas.kirsch@cs.ox.ac.uk \\
      \addr OATML, Department of Computer Science\\
      University of Oxford
}

\def\month{MM}  %
\def\year{YYYY} %
\def\openreview{\url{https://openreview.net/forum?id=XXXX}} %

\begin{document}

\maketitle

\begin{abstract}
The paper `Deep Learning on a Data Diet' by \citet{paul2021deep} introduces two innovative metrics for pruning datasets during the training of neural networks. While we are able to replicate the results for the EL2N score at epoch 20, the same cannot be said for the GraNd score at initialization. The GraNd scores later in training provide useful pruning signals, however.
The GraNd score at initialization calculates the average gradient norm of an input sample across multiple randomly initialized models before any training has taken place. Our analysis reveals a strong correlation between the GraNd score at initialization and the input norm of a sample, suggesting that the latter could have been a cheap new baseline for data pruning. Unfortunately, neither the GraNd score at initialization nor the input norm surpasses random pruning in performance. This contradicts one of the findings in \citet{paul2021deep}. We were unable to reproduce their CIFAR-10 results using both an updated version of the original JAX repository and in a newly implemented PyTorch codebase.
An investigation of the underlying JAX/FLAX code from 2021 surfaced a bug in the checkpoint restoring code that was fixed in April 2021\footnote{See \url{https://github.com/google/flax/commit/28fbd95500f4bf2f9924d2560062fa50e919b1a5}.}.
\end{abstract}

\section{Introduction}

The rapidly growing field of deep learning has led to significant advances in various applications, including computer vision, natural language processing, and speech recognition. However, the immense amounts of data required to train these neural networks present challenges in terms of computational resources, storage, and energy consumption. As a result, there is a growing interest in finding methods to reduce data requirements while still maintaining model performance. 

The senior author of `Deep Learning on a Data Diet' \citep{paul2021deep} recently gave a talk at our lab that explored this issue, presenting their novel metrics for pruning datasets. During the talk, the author of this current work suggested \emph{a correlation between the proposed GraNd score at initialization and input norms}, sparking further research into the effectiveness of these new pruning techniques. In this paper, we delve deeper into this intriguing question, exploring the practicality and efficacy of these metrics for data pruning.

\textbf{`Deep Learning on a Data Diet'.} %
\citet{paul2021deep} introduce two novel metrics: \emph{Error L2 Norm (EL2N)} and \emph{Gradient Norm at Initialization (GraNd)}. These metrics aim to provide a more effective means of dataset pruning. It is important to emphasize that the GraNd score at initialization is calculated before any training has taken place, averaging over several randomly initialized models. This fact has been met with skepticism by reviewers\footnote{See also \url{https://openreview.net/forum?id=Uj7pF-D-YvT&noteId=qwy3HouKSX}.}, but \citet{paul2021deep} specifically remark on GraNd at initialization:
\begin{quote}
\textbf{Pruning at initialization.} In all settings, GraNd scores can be used to select a training subset at initialization that achieves test accuracy significantly better than random, and in some cases, competitive with training on all the data. This is remarkable because GraNd only contains information about the gradient norm at initializion, averaged over initializations. This suggests that the geometry of the training distribution induced by a random network contains a surprising amount of information about the structure of the classification problem.
\end{quote}

\textbf{GraNd.} The GraNd score measures the magnitude of the gradient vector for a specific input sample in the context of neural network training over different parameter draws. The formula for calculating the (expected) gradient norm is:

\begin{equation}
\operatorname{GraNd}(x)=\mathbb{E}_{\theta_t}[\left\|\nabla_{\theta_t} L(f(x ; \theta_t), y)\right\|_2]
\end{equation}
 
where $\nabla_{\theta_t} L(f(x; \theta_t), y)$ is the gradient of the loss function $L$ with respect to the model's parameters $\theta_t$ at epoch $t$, $f(x; \theta)$ is the model's prediction for input $x$, and $y$ is the true label for the input. We take an expectation over several training runs.
The gradient norm provides information about the model's sensitivity to a particular input and helps in identifying data points that have a strong influence on the learning process.

\textbf{EL2N.} The EL2N score measures the squared difference between the predicted and (one-hot) true labels for a specific input sample. The formula for calculating the EL2N score is:

\begin{equation}
\operatorname{EL2N}(x)=\mathbb{E}_{\theta_t}[\|f(x ; \theta_t)-y\|_2^2]
\end{equation}
 
where $f(x; \theta)$ is the model's prediction for input $x$, $y$ is the (one-hot) true label for the input, and $\lVert \cdot \rVert_2$ denotes the Euclidean (L2) norm. The EL2N score provides insight into the model's performance on individual data points, allowing for a more targeted analysis of errors and potential improvements.

The GraNd and EL2N scores are proposed in the context of dataset pruning, where the goal is to remove less informative samples from the training data. Thus, one can create a smaller, more efficient dataset that maintains the model's overall performance while reducing training time and computational resources.

While GraNd at initialization does not require model training, it requires a model and is not cheap to compute. In contrast, the input norm of training samples is incredibly cheap to compute and would thus provide an exciting new baseline to use for data pruning experiments. We investigate this correlation in this paper and find positive evidence for it. However, we also find that the GraNd score at initialization does not outperform random pruning, unlike the respective results of \citet{paul2021deep} for GraNd at initialization.

\begin{figure}
    \centering
    \begin{subfigure}[b]{0.5\textwidth}
        \centering
        \includegraphics[width=\textwidth]{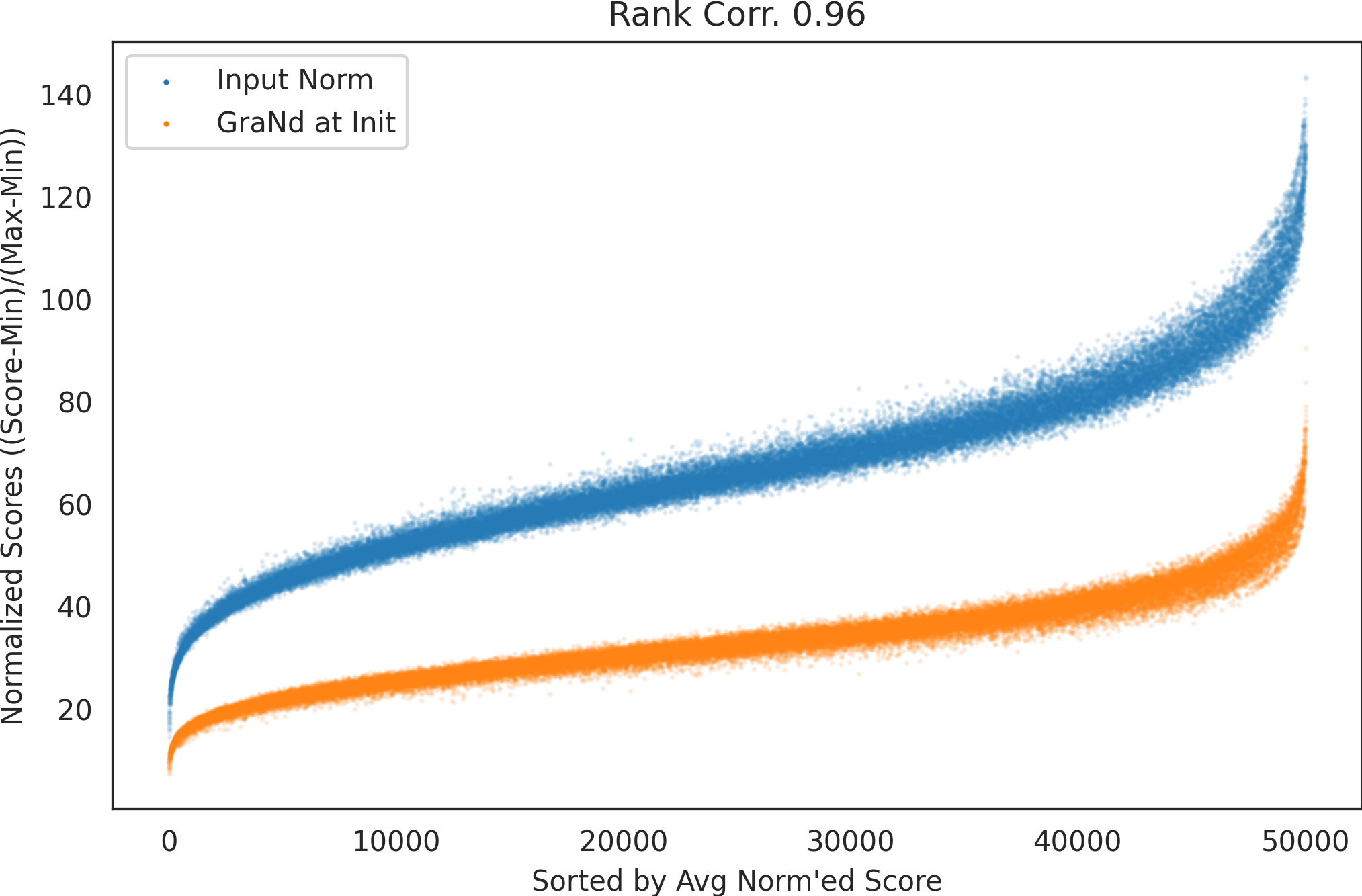}
        \caption{Original Repo (10 trials)}
        \label{fig:grand_init_vs_input_norm:original}
    \end{subfigure}\hfill
    \begin{subfigure}[b]{0.5\textwidth}
        \centering
        \includegraphics[width=\textwidth]{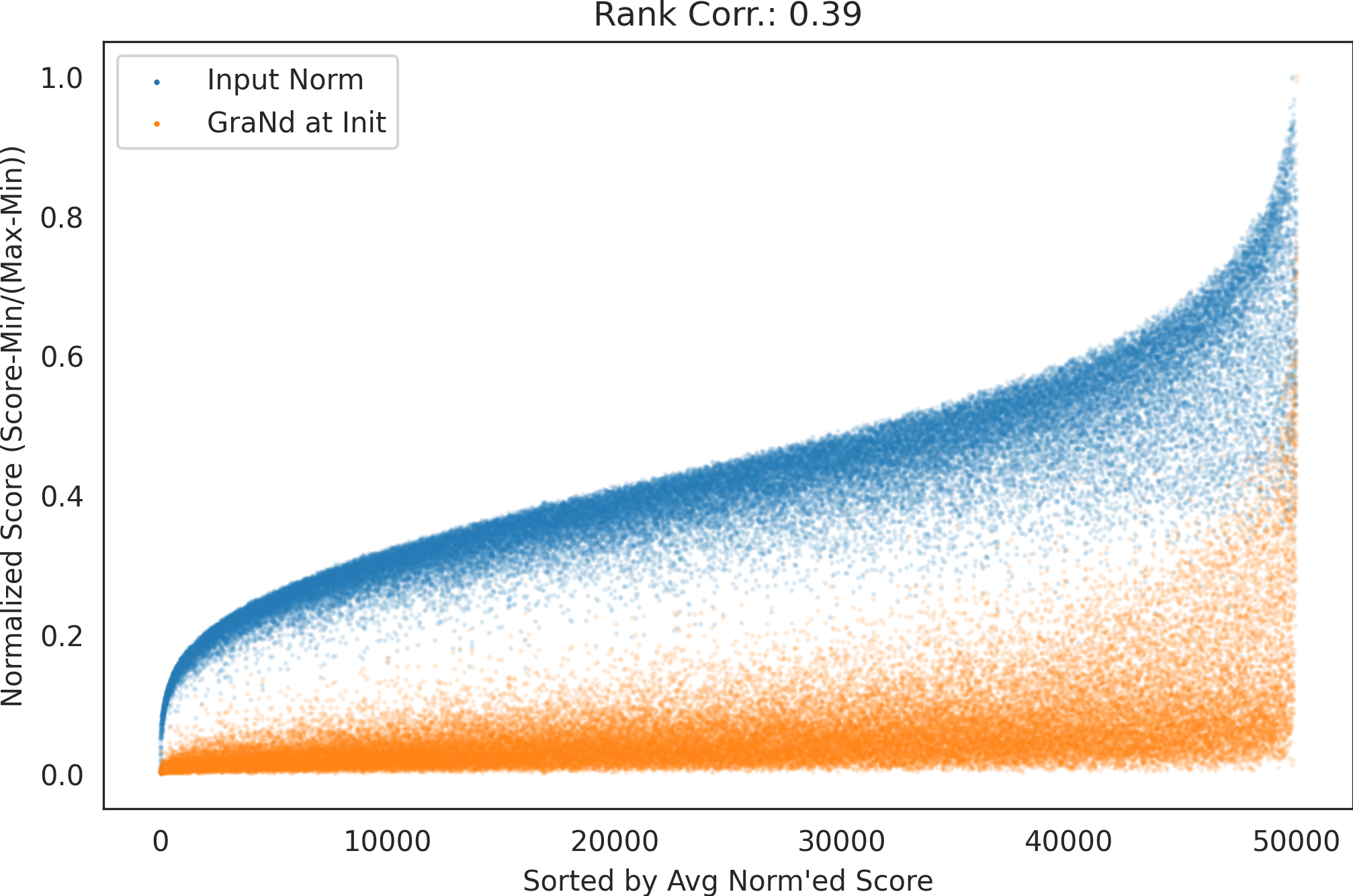}
        \caption{Hlb (10 trials)}
        \label{fig:grand_init_vs_input_norm:hlb}
    \end{subfigure}
    \begin{subfigure}[b]{0.5\textwidth}
        \centering
        \includegraphics[width=\textwidth]{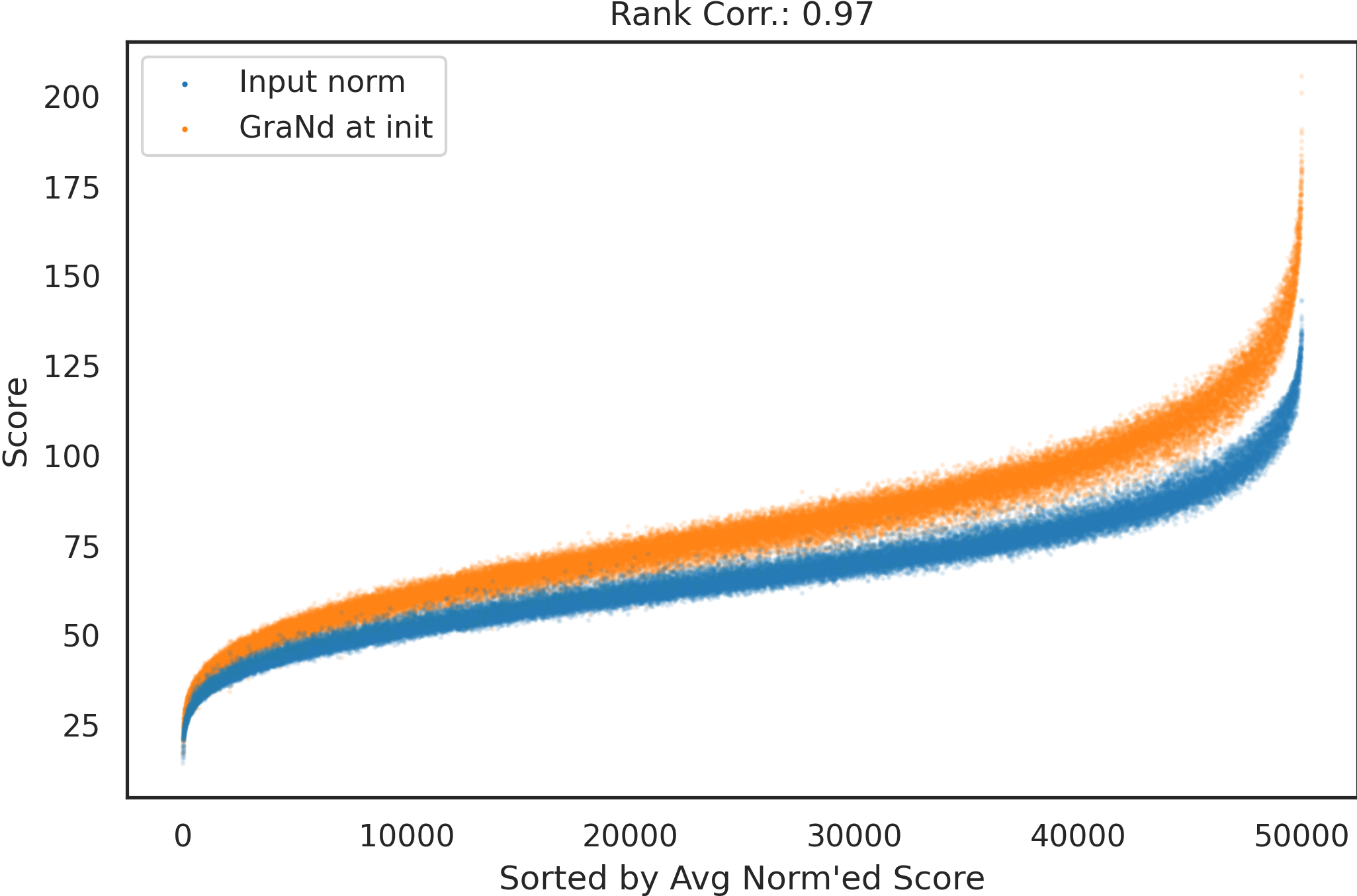}
        \caption{Minimal (120 trials)}
        \label{fig:grand_init_vs_input_norm:joost}
    \end{subfigure}\hfill
    \begin{subfigure}[b]{0.5\textwidth}
      \centering
      \includegraphics[width=\textwidth]{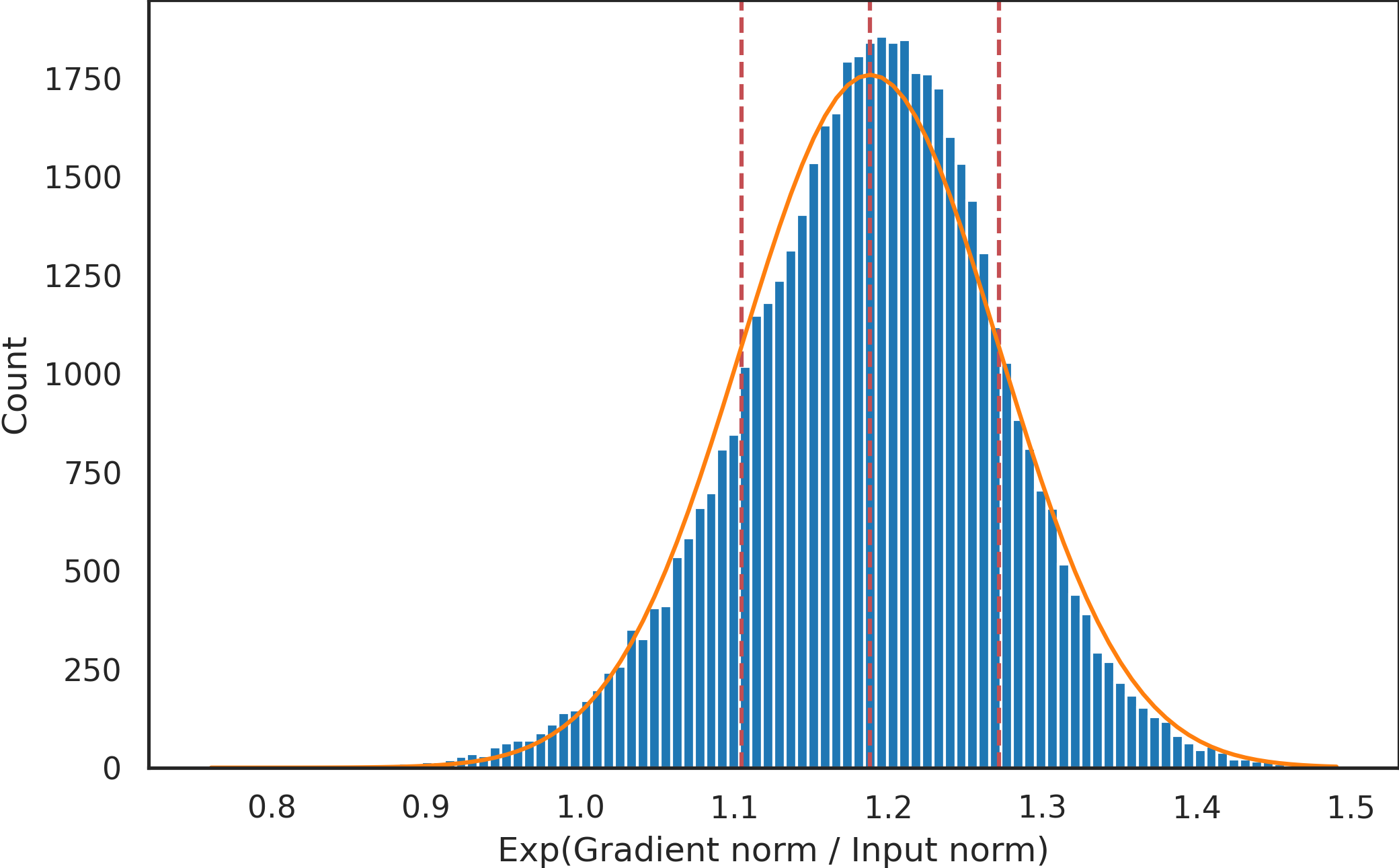}
      \caption{Minimal (120 trials)}
      \label{fig:grand_init_div_input_norm:joost}
  \end{subfigure}
    \caption{\emph{Correlation between GraNd at Initialization and Input Norm for CIFAR-10's training set.}
    \textbf{(\subref{fig:grand_init_vs_input_norm:original}, \subref{fig:grand_init_vs_input_norm:hlb}, \subref{fig:grand_init_vs_input_norm:joost})}: We sort the samples by their average normalized score (i.e., the score minus its minimum divided by its range), plot the scores and compute Spearman's rank correlation on CIFAR-10's training data. The original repository and the `minimal' implementation have very high rank correlation---`hlb' has a lower but still strong rank correlation.
    \textbf{(\subref{fig:grand_init_div_input_norm:joost})}: \emph{Ratio between input norm and gradient norm.} In the `minimal' implementation, the ratio between input norm and gradient norm is roughly log-normal distributed.}
    \label{fig:grand_init_vs_input_norm}
\end{figure}

\textbf{Outline.} %
In \S\ref{sec:correlation}, we begin by discussing the correlation between input norm and gradient norm at initialization.
We empirically find strong correlation between GraNd scores at initialization and input norms as we average over models.
In \S\ref{subsec:repro_fig1}, we explore the implication of this insight for dataset pruning and find that both GraNd at initialization and input norm scores do not outperform random pruning, but GraNd scores after a few epochs perform similar to EL2N scores at these later epochs.

In summary, this reproduction contributes a new insight on the relationship between input norm and gradient norm at initialization and finds a failure to reproduce one of the six contributions of \citet{paul2021deep}.

\section{Investigation}
\label{sec:reproduction}

\begin{figure}
  \centering
  \begin{subfigure}[b]{0.5\textwidth}
      \centering
      \includegraphics[width=\textwidth]{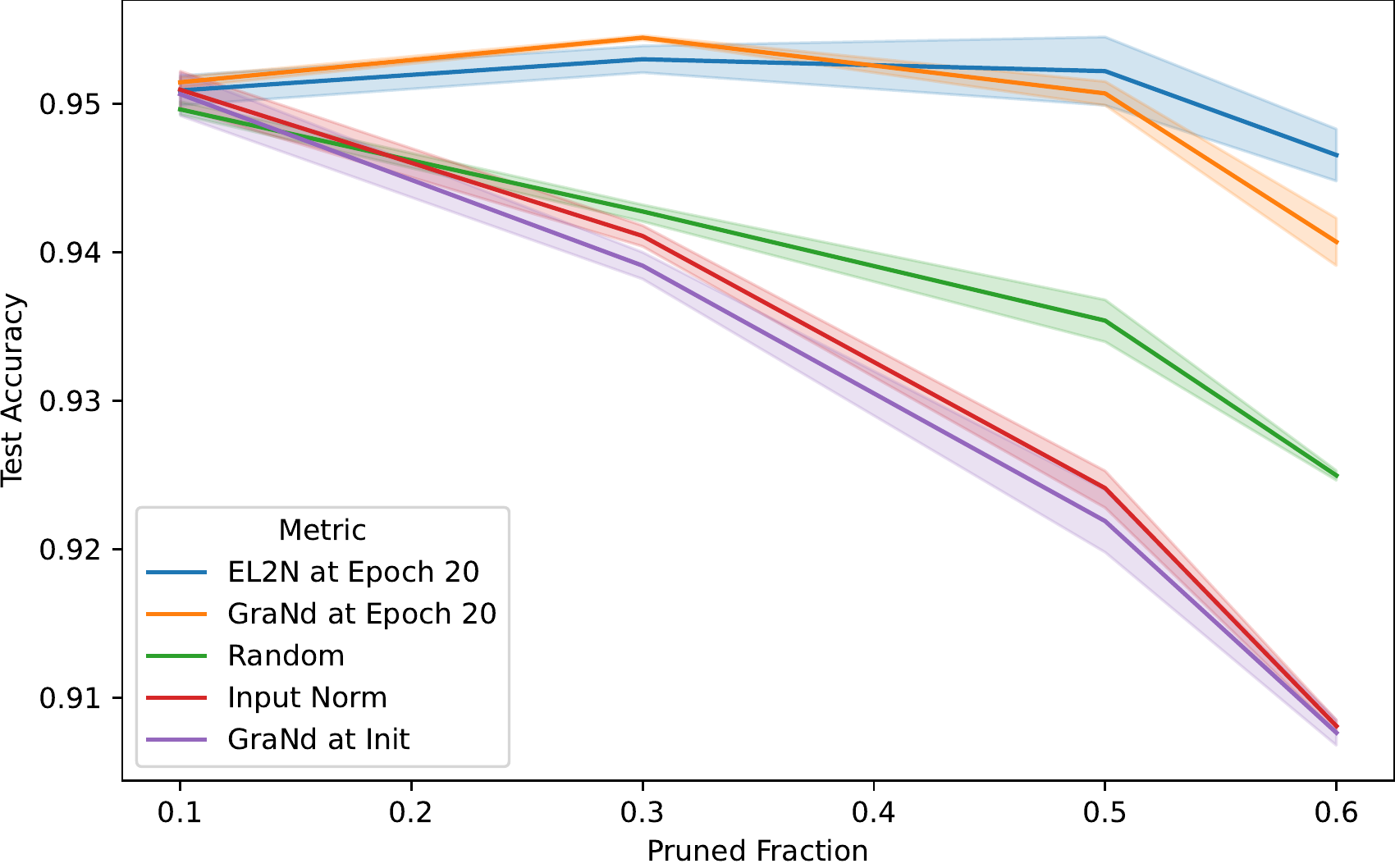}
      \caption{Original Repo (2 trials each)}
      \label{fig:fig1_repro:original}
  \end{subfigure}\hfill
  \begin{subfigure}[b]{0.5\textwidth}
    \centering
    \includegraphics[width=\textwidth]{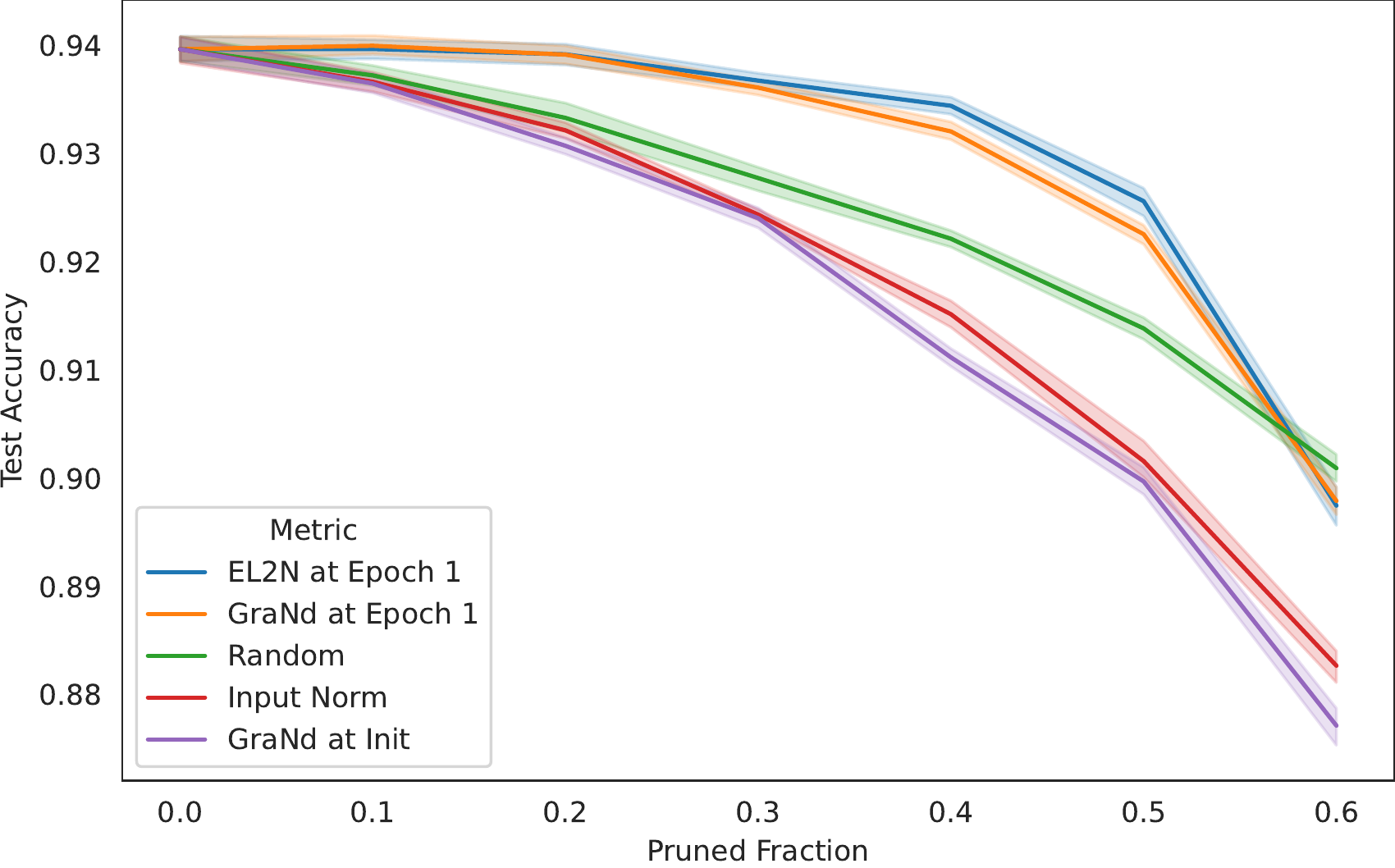}
    \caption{Hlb (10 trials)}
    \label{fig:fig1_repro:hlb}
  \end{subfigure}
  \caption{\emph{Reproduction of Figure 1 (second row) from \citet{paul2021deep}.}
  In both reproductions, GraNd at initialization performs as well as the input norm. However, it does not perform better than random pruning. Importantly, it also fails to reproduce the results from \citet{paul2021deep}. However, GraNd at epoch 20 (respectively at epoch 1 for `hlb') performs similar to EL2N and like GraNd at initialization in \citet{paul2021deep}.}
  \label{fig:fig1_repro}
\end{figure}

We investigate the correlation between input norm and GraNd at initialization and the other scores on CIFAR-10 \citep{krizhevsky2009learning} in three different ways:
First, we update the original paper repository\footnote{
  \url{https://github.com/blackhc/data_diet}
} (\url{https://github.com/mansheej/data_diet}), which uses JAX \citep{jax2018github}, rerun the experiments for Figure 1 (second row) in \citet{paul2021deep} for CIFAR-10, which trains for 200 epochs, using GraNd at initialization, GraNd at epoch 20, E2LN at epoch 20, Forget Score at epoch 200, and input norm. Second, we reproduce the same experiments using `hlb' \citep{Balsam_hlb-CIFAR10_2023}, which is a strongly modified version of ResNet-18 that allows to train to high accuracy in 12 epochs taking about 30 seconds total on a Nvidia RTX 4090 in PyTorch \citep{pytorchgithub} For the latter, we compare GraNd at initialization, GraNd at epoch 1 ($\approx 20/200\cdot 12$ epochs), EL2N at epoch 1, and input norm\footnote{\label{code:pytorch}\url{https://github.com/blackhc/pytorch_datadiet}}.
Third, we compare the rank correlations between the different scores for those two repositories and also use another `minimal' CIFAR-10 implementation \citep{Joost_minimal_cifar10} with a standard ResNet18 architecture for CIFAR-10 to compare the rank correlations.

\subsection{Correlation between GraNd at initialization and input norm}
\label{sec:correlation}

To better understand the relationship between the input norm and the gradient norm at initialization, let us consider a toy example first and then appeal to empirical evidence as is common in deep learning research:
let's examine linear softmax classification with $C$ classes (without a bias term). The model takes the form:
\begin{equation}
  f(x) = \operatorname{softmax}(W x),
\end{equation}
together with the cross-entropy loss function:
\begin{equation}
L = -\log f(x)_y.
\end{equation}
The gradient of the loss function with respect to the rows $w_j$ of the weight matrix $W$ is:
\begin{equation}
\nabla_{w_j} L = (f(x)_j - \mathbbm{1}\{j = y\}) x
\end{equation}
where $\mathbbm{1}\{j = y\}$ is the indicator function that is 1 if $j = y$ and 0 otherwise.
The squared norm of the gradient is:
\begin{equation}
\| \nabla_w L \|_2^2 = \sum_{j = 1}^C (f(x)_j - \mathbbm{1}\{j = y\})^2 \|x\|_2^2.
\end{equation}
In expectation over $W$ (different initializations), the norm of the gradient is:
\begin{equation}
\mathbb{E}_W \left [\| \nabla_w L \|_2 \right ] = \mathbb{E}_W \left [\left (\sum_{j = 1}^C (f(x)_j - \mathbbm{1}\{j = y\})^2 \right )^{1/2} \right ] \|x\|_2.
\end{equation}
Thus, we see that the gradient norm is a multiple of the input norm. The factor depends on $f(x)_j$, which we could typically expect to be $1/C$.

\textbf{Empirical Evidence.} %
In \Cref{fig:grand_init_vs_input_norm}, we see that on CIFAR-10's training set, GraNd at initialization and the input norm are highly correlated. This is true for the original repository, the `hlb' and the `minimal' implementation. The `hlb' implementation has a lower but still strong correlation.

\subsection{Reproducing Figure 1 of \citet{paul2021deep} on CIFAR-10}
\label{subsec:repro_fig1}

\begin{figure}
  \centering
  \begin{subfigure}[b]{0.5\textwidth}
      \centering
      \includegraphics[width=\textwidth]{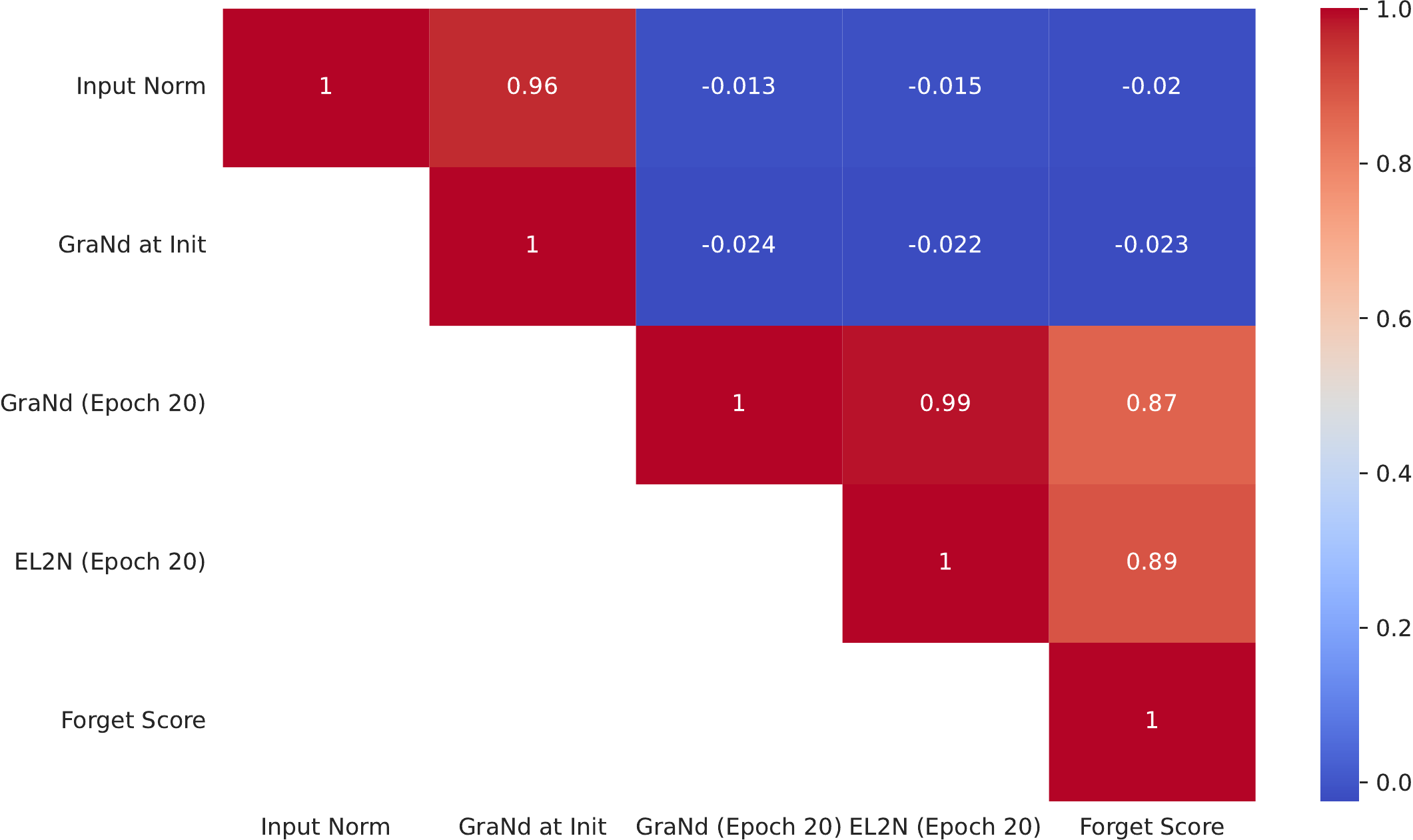}
      \caption{Original Repo (2 trials each)}
      \label{fig:heatmap:original}
  \end{subfigure}\hfill
  \begin{subfigure}[b]{0.5\textwidth}
    \centering
    \includegraphics[width=\textwidth]{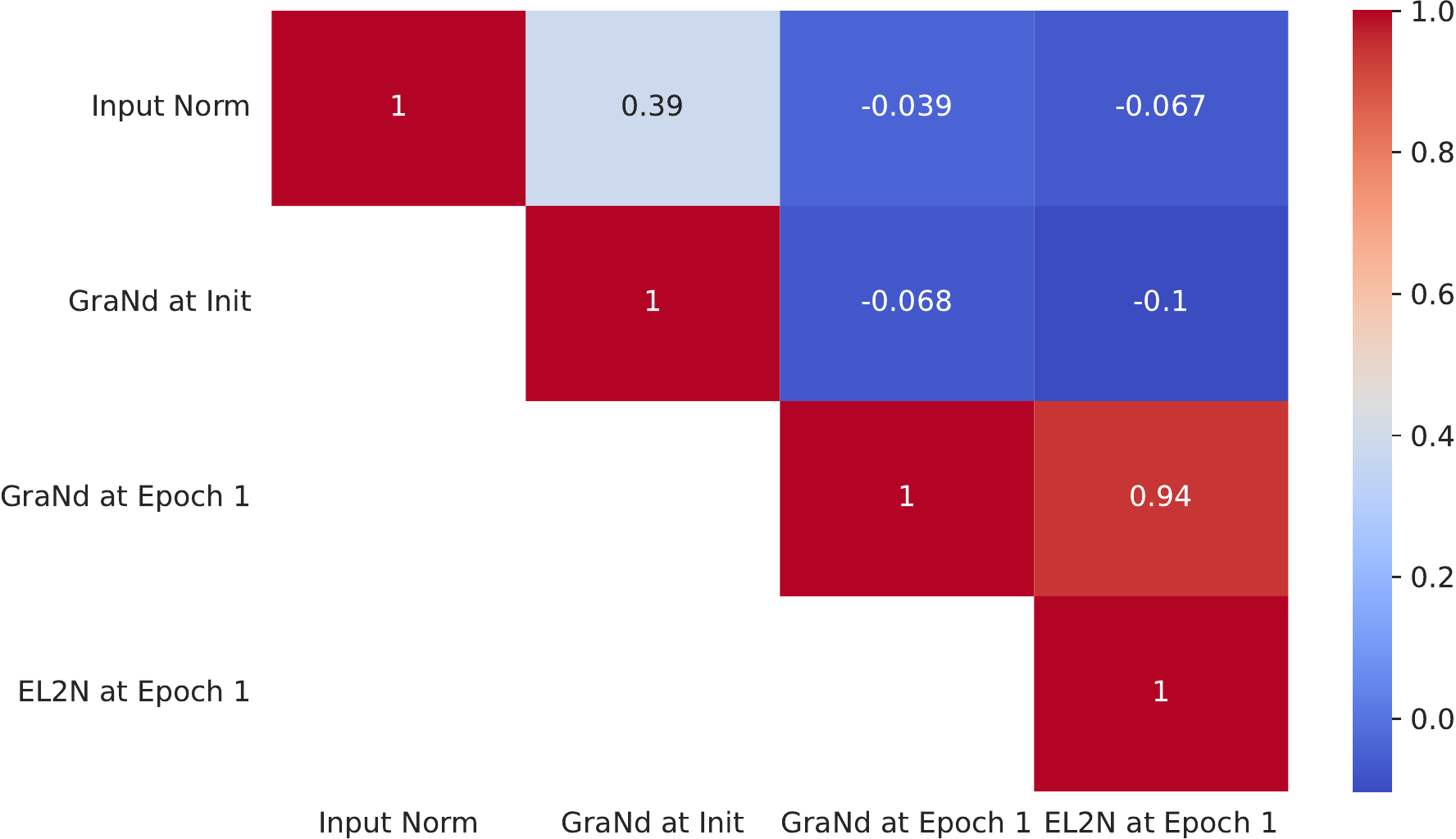}
    \caption{Hlb (10 trials)}
    \label{fig:heatmap:hlb}
  \end{subfigure}
  \caption{\emph{Rank Correlations of the Scores.} Cf. Figure 12 in the appendix of \citet{paul2021deep}.
  In both reproductions, GraNd at initialization and input norm are positively correlated, while GraNd and EL2N at later epochs are strongly correlated with each other and the Forget Score (at epoch 200).}
  \label{fig:heatmap}
\end{figure}

In \Cref{fig:fig1_repro}, we see that GraNd at initialization performs about as well as using the input norm. However, it does not reproduce the results from \citet{paul2021deep}.
It performs worse than random pruning (for `hlb'). However, GraNd at epoch 20 (respectively at epoch 1 for `hlb') performs like GraNd at initialization in \citet{paul2021deep}.
Similarly, in \Cref{fig:heatmap}, we see that GraNd at initialization and the input norm are strongly correlated as are GraNd at later epochs, EL2N and the Forget Score, with little correlation between these two groups.

\section{Conclusion}

If GraNd at initialization performed as well as claimed in \citet{paul2021deep}, using the input norm would provide a new exciting baseline for data pruning because it is model independent and cheaper to compute than GraNd or other scores.
However, since only GraNd at later epochs seems to perform as expected, we cannot recommend using input norm or GraNd at initialization for data pruning.

As to the failure to reproduce the results of \citet{paul2021deep}, we could not rerun the code using the original JAX version because it is too old for our GPU. The authors of \citet{paul2021deep} were, however, able to set up a Google Cloud VM with an old image that was able to reproduce the original results using the original JAX version. On further investigation, the author of this reproduction found a bug in \texttt{flax.training.restore\_checkpoint} that was fixed in April 2021\footnote{See \url{https://github.com/google/flax/commit/28fbd95500f4bf2f9924d2560062fa50e919b1a5}.}: passing a \texttt{0} step (i.e. initialization) would trigger loading the \emph{latest} checkpoint instead of the zero-th checkpoint because the internal implementation was checking 
\texttt{if step:} instead of \texttt{if step is not None:} when deciding whether to fallback to loading the latest checkpoint. This bug was fixed in April 2021, but the authors of \citet{paul2021deep} were not aware of this bug and did not rerun their experiments with newer JAX/FLAX versions. We have accordingly informed the authors of \citet{paul2021deep}.

\subsubsection*{Acknowledgments}
Thanks to Mansheej Paul and Karolina Dziugaite for very helpful feedback and discussions.
AK is supported by the UK EPSRC CDT in Autonomous Intelligent Machines and Systems (grant reference EP/L015897/1). ChatGPT was used to suggest edits.

\bibliography{main}
\bibliographystyle{tmlr}

\newpage

\appendix
\section{Appendix}

\begin{figure}[h]
  \centering
  \begin{subfigure}[b]{0.5\textwidth}
      \centering
      \includegraphics[width=\textwidth]{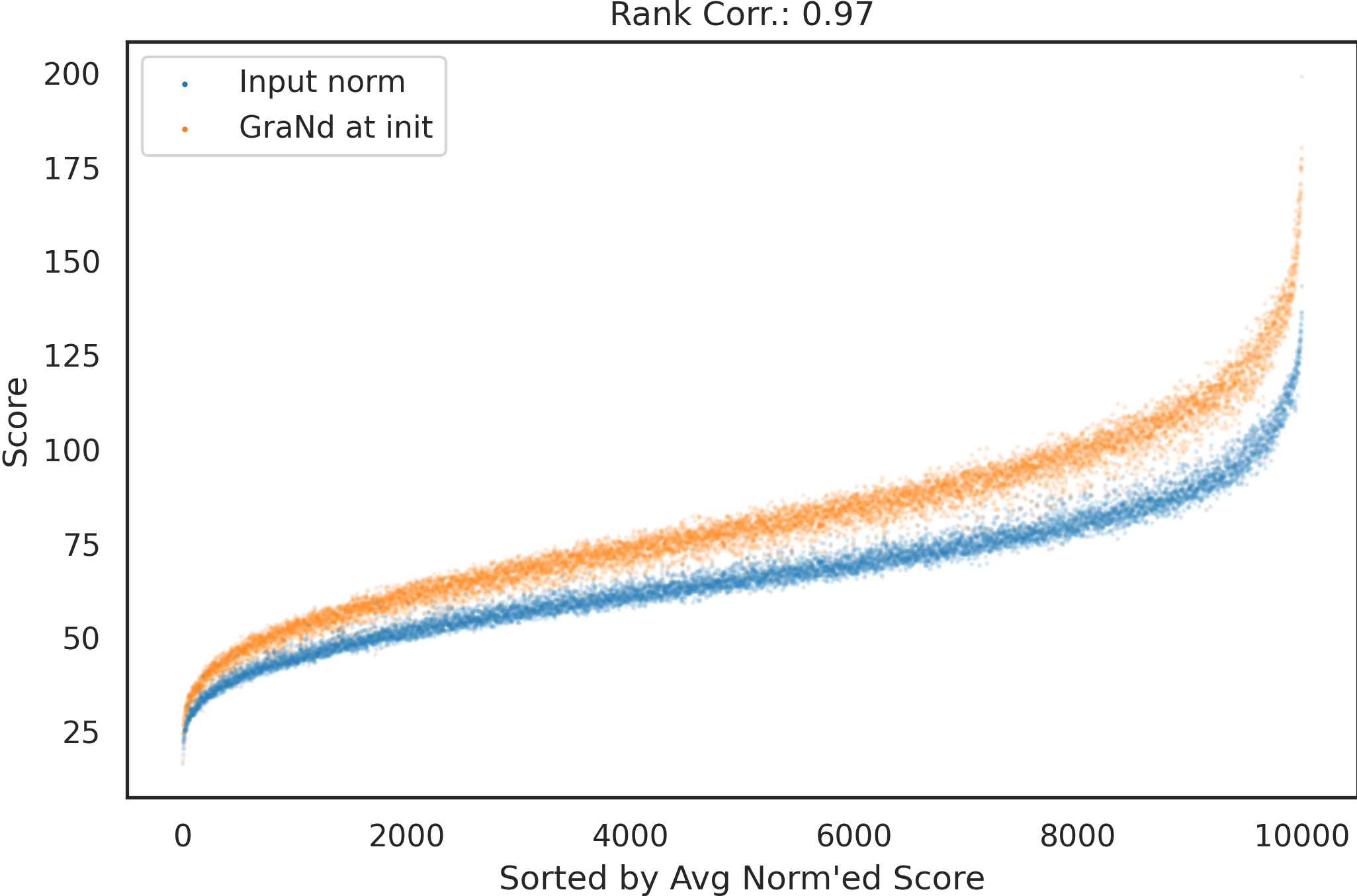}
      \caption{Minimal (1000 trials)}
      \label{fig:test_grand_init_vs_input_norm:joost}
  \end{subfigure}\hfill
  \begin{subfigure}[b]{0.5\textwidth}
    \centering
    \includegraphics[width=\textwidth]{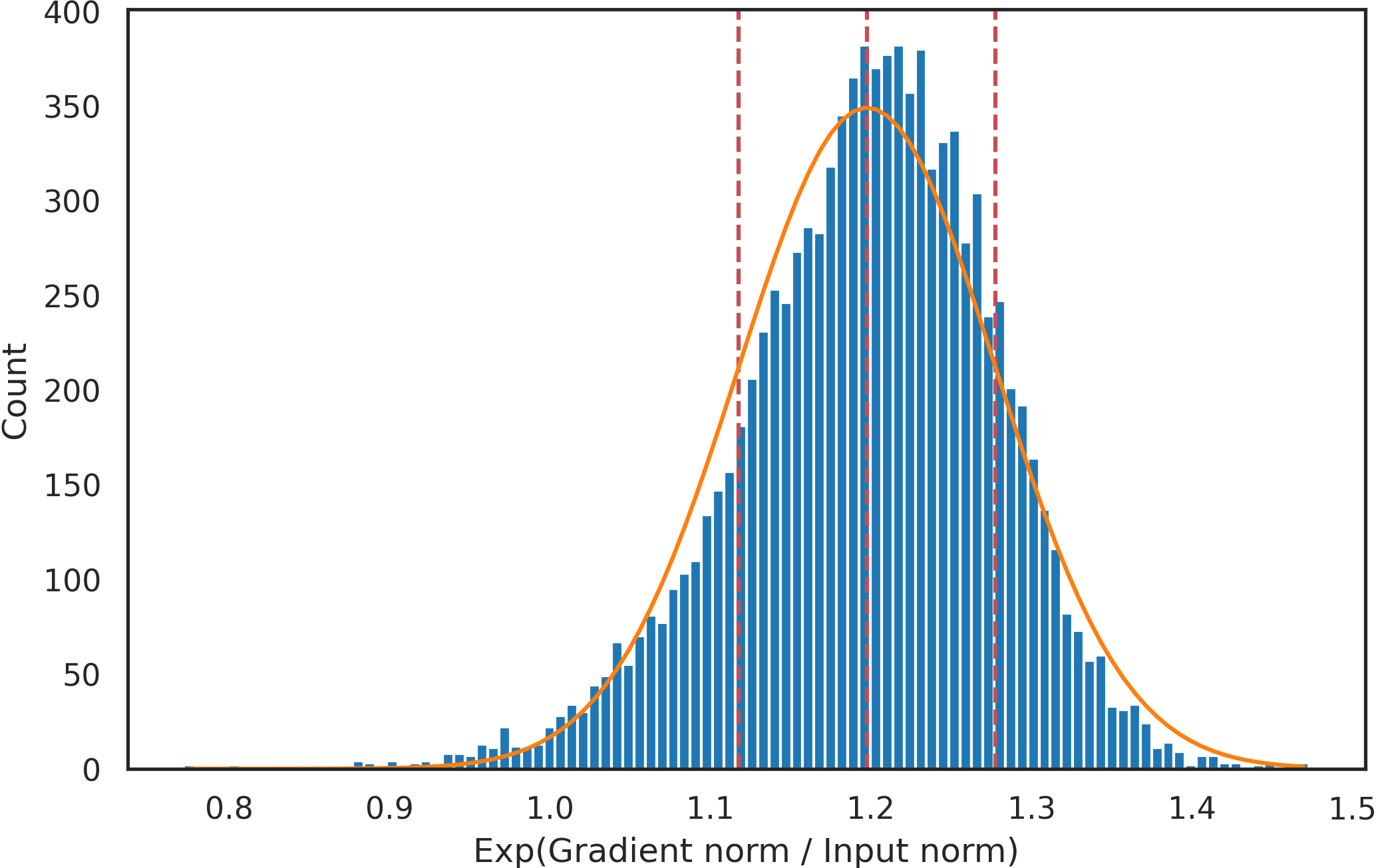}
    \caption{Minimal (1000 trials)}
    \label{fig:test_grand_init_div_input_norm:joost}
\end{subfigure}
  \caption{\emph{Correlation between GraNd at Initialization and Input Norm on the Test Set.}
  \textbf{(\subref{fig:test_grand_init_vs_input_norm:joost})}: We sort the samples by their average normalized score (i.e., the score minus its minimum divided by its range), plot the scores and compute Spearman's rank correlation on CIFAR-10's test data. The original repository and the `minimal' implementation have very high rank correlation---`hlb' has a lower but still strong rank correlation.
  \textbf{(\subref{fig:test_grand_init_div_input_norm:joost})}: \emph{Ratio between input norm and gradient norm.} In the `minimal' implementation, the ratio between input norm and gradient norm is roughly log-normal distributed}
  \label{fig:test_grand_init_vs_input_norm}
\end{figure}

\end{document}